\documentclass[a4paper,fleqn]{cas-dc}

\usepackage{soul, color, xcolor}
\usepackage{color,array,amsmath}
\usepackage{graphicx}
\usepackage{overpic}
\usepackage{graphicx}
\usepackage{multirow}
\usepackage{booktabs}
\usepackage{amsmath}
\usepackage{float}  
\usepackage{lipsum}
\usepackage{lineno,hyperref}
\usepackage[round]{natbib}
\usepackage[ruled]{algorithm2e}
\usepackage{algpseudocode}
\usepackage{multirow}
\usepackage{booktabs}
\usepackage{color, soul} 
\usepackage {hyperref}
\usepackage{wrapfig}
\usepackage{orcidlink}


\begin{document}
\let\printorcid\relax
\title {\Large Similarity-based Context Aware Continual Learning for Spiking Neural Networks}
\shorttitle {Similarity-based context aware continual learning for spiking neural networks}
\shortauthors{B.Han, F.Zhao, Y.Li, Q.Kong, X.Li and Y.Zeng}

\author[1,3,6]{Bing Han \orcidlink{0000-0003-2424-7086}}

\author[1,6]{Feifei Zhao \orcidlink{0000 0002 4156 2750}}

\author[1,3]{Yang Li}

\author[1,2,3]{Qingqun Kong}

\author[3]{Xianqi Li}

\author[1,2,3,4,5]{Yi Zeng \orcidlink{0000-0002-9595-9091} \corref{cor1}}

\address[1]{Brain-inspired Cognitive Intelligence Lab, Institute of Automation, Chinese Academy of Sciences, Beijing, China.}
\address[2]{School of Future Technology, University of Chinese Academy of Sciences, Beijing, China.}
\address[3]{School of Artificial Intelligence, University of Chinese Academy of Sciences, Beijing, China.}
\address[4]{Key Laboratory of Brain Cognition and Brain-inspired Intelligence Technology, Chinese Academy of Sciences, Shanghai, China.}
\address[5]{Center for Long-term Artificial Intelligence, Beijing, China.}
\address[6]{These authors contributed equally to this work.}
\cortext[cor1]{Corresponding author: yi.zeng@ia.ac.cn, 100190}

\begin{abstract}
Biological brains have the capability to adaptively coordinate relevant neuronal populations based on the task context to learn continuously changing tasks in real-world environments. However, existing spiking neural network-based continual learning algorithms treat each task equally, ignoring the guiding role of different task similarity associations for network learning, which limits knowledge utilization efficiency. Inspired by the context-dependent plasticity mechanism of the brain, we propose a Similarity-based Context Aware Spiking Neural Network (SCA-SNN) continual learning algorithm to efficiently accomplish task incremental learning and class incremental learning. Based on contextual similarity across tasks, the SCA-SNN model can adaptively reuse neurons from previous tasks that are beneficial for new tasks (the more similar, the more neurons are reused) and flexibly expand new neurons for the new task (the more similar, the fewer neurons are expanded). Selective reuse and discriminative expansion significantly improve the utilization of previous knowledge and reduce energy consumption. Extensive experimental results on CIFAR100, ImageNet generalized datasets, and FMNIST-MNIST, SVHN-CIFAR100 mixed datasets show that our SCA-SNN model achieves superior performance compared to both SNN-based and DNN-based continual learning algorithms. Additionally, our algorithm has the capability to adaptively select similar groups of neurons for related tasks, offering a promising approach to enhancing the biological interpretability of efficient continual learning.
\end{abstract}


\maketitle
\section{Introduction}
Lifelong learning is the prominent capability of biological intelligence and the significant challenge of artificial intelligence. In the process of continuously encountering new environments, the brain effectively identifies the connections between new and old knowledge through task contexts. It reshapes the neural network by associating new tasks with similar prior knowledge to adapt to new information, while strengthening the old knowledge~\cite{bar2007proactive,bar2004visual}. However, there is still much room for existing continual learning research to improve context-based efficient and flexible learning.

Spiking Neural Networks (SNNs)~\mbox{\cite{Maass1997Networks}} have been extensively researched due to their high efficiency and bio-interpretability, incorporating SNNs with continual learning mechanisms of the brain provides natural advances. Nevertheless, to the best of our knowledge, there is little continual learning for SNNs. Only ASP~\mbox{\cite{panda2017asp}} and HMN~\mbox{\cite{zhao2022framework}} use STDP-based regularization and neuronal activity-based subnetwork selection to overcome catastrophic forgetting, but they are only suitable for shallow networks. DSD-SNN~\mbox{\cite{han2023enhancing}} and SOR-SNN~\mbox{\cite{han2023adaptive}} apply brain-inspired continual learning algorithms to deep SNNs, ignoring the effect of task-to-task associations.

The brain adaptively modulates neuronal generation, allocation, extinction, and reuse for continual learning. Therefore, except for a few continual learning algorithms based on SNNs, we also focus on structure extension algorithms based on DNNs~\mbox{\cite{van2019three}}. Structure expansion methods  assign separate sub-network structures for different tasks, ensuring that learning new tasks does not interfere with previous ones. They can be categorized into progressive neural networks and subnetwork selection algorithms. Among them, progressive networks require assigning a new network to each new task, with full connectivity between the old and new networks~\cite{rusu2016progressive,siddiqui2021progressive}. Subnetwork selection algorithms select task-specific sparse masks among a finite network~\mbox{\cite{fernando2017pathnet,gao2022efficient}} ~\mbox{\cite{chandra2023continual,hutask2024task}}.

However, these two structure expansion algorithms face the following challenges: 1) Catastrophic increase in energy consumption due to the growth of the network scale. The number of progressive neural network parameters increases linearly with the number of tasks~\cite{yan2021dynamically,huang2023resolving}. Despite the subnetwork selection algorithms managing network size, the global network state transitions from  sparse to fully connected configuration, resulting in augmented energy consumption~\cite{rajasegaran2019random,xu2018reinforced}. 2) Cross-task knowledge transfer: the progressive network reuses all past knowledge indiscriminately~\cite{rusu2016progressive}, and the subnetwork algorithms do not take into account the association between tasks when learning task-specific masks~\cite{dekhovich2022continual,sokar2021spacenet}. 3) Structure expansion algorithms rely on prior knowledge of the task affiliation of the current sample to determine the relevant elements to utilize~\cite{yoon2017lifelong,chandra2023continual}. Thus, most algorithms are primarily suited for task-incremental learning (completing a given task in testing) and are insufficient for class-incremental learning (completing all learned tasks in testing).

To address these issues, DNN-based continual learning has been drawn to some extent from the biological continual learning mechanisms~\mbox{\cite{ma2023dual}}, such as contextual similarity recognition. They consider similarity usually focus on the synapses~\mbox{~\cite{ke2022continual}} and data~\mbox{~\cite{wang2023task}} and involve designing additional evaluation networks~\mbox{~\cite{ke2020continual}}. Upon identifying similar tasks, all neurons of the similar tasks are directly reused without any distinction ~\cite{wang2022toward}. These factors can lead to different parts in similar tasks interfering, neither are they able to activate similar neural circuits in similar tasks as the brain does.

Inspired by the brain contextual task knowledge recognition and flexible neural circuit allocate and reuse mechanisms, we propose the Similarity-based Context Aware continual learning for Spiking Neural Networks (SCA-SNN). Firstly, we designed a task similarity evaluation method that integrates the current data and network state to determine task similarity. Leveraging this assessment, we adaptively reuse neurons from previous tasks and generate a certain number. The underlying principle is that the more similar the new task is to the previous one, the less new network expansion and the more reuse of existing network. More importantly, to avoid redundancy caused by the extensive reuse of neurons, we draw inspiration from the developmental plasticity of human brain neurons, which adhere to the principle of 'use it or lose it' ~\cite{bruer1999neural}. This means that neurons habituated to certain tasks require stronger repetitive stimulation to be reactivated in a new task~\cite{grissom2009habituation}, while those unrelated to the new tasks will be disconnected. We design a gradient-based method for selecting reused neurons, ensuring that only neurons contributing effectively to the new task are reused.

We validate the proposed model on CIFAR100, mini-Imagenet general continual learning dataset, and FMNIST-MNIST, SVHN-CIFAR100 mixed dataset. Our model effectively identifies task similarity relationships to guide more flexible and efficient neuron allocation, thereby reducing energy consumption and achieving superior performance. Meanwhile, the proposed model can adaptively assign similar neuron populations to similar tasks like the human brain. Our SCA-SNN contribution points are as follows: 

\begin{enumerate}
	\item[$\bullet$]We propose the Similarity-based Context Aware model in SNN, which leverages task similarity to selectively reuse similar neurons from previous tasks and flexibly expand new neurons for new tasks. The proposed model promotes cross-task knowledge transfer and belongs to advanced exploration for  brain-inspired SNN continual learning.
	\item[$\bullet$] We design an effective neuron reuse strategy inspired by the habituation mechanism observed in the biological brain neurons. This approach carefully selects neurons that are truly beneficial for new tasks. By filtering out superfluous neurons,  our approach enhances the efficiency of energy and utilization of known knowledge, preventing irrelevant neurons from interfering with the execution of new tasks.
	\item[$\bullet$] Extensive experiments across various class-incremental and task-incremental learning demonstrate that the proposed model achieves the state-of-the-art performance of the spiking neural networks. Meanwhile, due to the sparsity of the network and the discrete characterization of SNN, the proposed method significantly reduces the energy consumption. 

\end{enumerate}

\section{Related Works}

\textbf{Structure expansion continual learning.} Continual learning algorithms based on structural expansion separate the features of old and new tasks and prepare a unique set of parameters for each task. The most straightforward approach is to construct a separate network for each task. To integrate knowledge,  progressive neural network (PNN) ~\mbox{\cite{rusu2016progressive,siddiqui2021progressive}} adds lateral connections between old and new task networks and freezes the old task network. DER~\cite{yan2021dynamically} adds an attention layer to synthesize the convolutional features of all network outputs and combines with iCaRL~\cite{rebuffi2017icarl} algorithm achieving class incremental learning.  However, as the number of tasks increases, the parameters of the PNN expand rapidly. To address this issue, sub-network selection algorithms have been proposed to identify a sub-network for each task within a fixed-size network. DEN~\mbox{\cite{yoon2017lifelong}}, SupSup
~\mbox{\cite{wortsman2020supermasks}} and TAME~\mbox{\cite{zhu2024tame}} use pruning algorithms and PathNet~\mbox{\cite{yoon2017lifelong}} use genetic algorithms to select sub-networks inspirationally. RCL~\cite{xu2018reinforced} and CLEAS~\cite{gao2022efficient} use an auxiliary reinforcement learning network while HNET~\cite{von2019continual} and CHT~\cite{vladymyrov2023continual} uses an auxiliary hypernetwork to select task-specific subnetworks in the main network. However, these structure expansion algorithms treat each task equally and do not consider the association between tasks, limiting knowledge transfer efficiency.

\textbf{Similarity contexts in continual learning.}  The brain is adept at utilizing old knowledge to enhance learning new tasks. Hence, recent research is concerned with recognizing the similarity between tasks to guide learning new tasks. CAT~\cite{ke2020continual} first trains a continual learning model in a series of similar and dissimilar mixed tasks and proposes a similarity assessment method based on statistical risk. Specifically, when the new task performance trained based on the old task network outperforms that trained with random initialization, CAT considers the new task similar to the old task. PAR~\cite{wang2023task} uses Kullback-Leibler scatter to assess the distributional similarity of the input data but ignores the effect of the existing network state on task similarity. Both methods discussed above require additional similarity assessment networks. SDR~\mbox{\cite{wang2022toward}} utilizes the Dirichlet Process Mixture Model (DPMM) to evaluate the distributional similarity of data within the current network. These methods reuse all the parameters of the old task after determining that the tasks are similar, but even similar tasks often have a subset of irrelevant parameters that interfere with the learning of the new task. Overall, most of the existing continual learning based on structure expansion and context similarity belongs to DNNs,  little to SNNs. This paper aims to explore context-based continual learning specialized for SNN to improve knowledge and energy efficiency.


\section{Prerequisites}
\subsection{Continual Learning Problem Setting}
In continual learning, the model sequentially learns a series of non-smooth tasks $\{T_1,...,T_t,...,T_N\}$. Each task $T_t$ has a training set $D_{train}^t=\{(x_i,y_i), i=1,...,n_{train}^t\}$ where $x_i$ is the picture data, $y_i$ is the true lable and $n_{train}^t$ is the number of training samples. Similarly, $T_t$ has a testing set $D_{test}^t$. When the task $T_t$ comes, the model already has the structure and function of the past learned tasks $\{T_1,...,T_{t-1}\}$. During training, we continually train based on the existing model using $D_{train}^t$. The $\{D_{train}^1$,..., $D_{train}^{t-1}\}$ are overwhelmingly invisible. Only a few samples can be saved in the memory store. In testing, task incremental learning (TIL) assumes that the task identifiers of the current samples are known, i.e., separately test $\{D_{test}^1$,..., $D_{test}^{t}\}$ as follow:
\begin{equation}
    \label{t}
    \mathop{max}\limits_{\theta} \sum_{i}^{T_t} E_{(x_j,y_j)\sim T_i}[log p_{\theta}(y_j|x_j,T_i)]
\end{equation}
where $\theta$ is the network parameters. However, in the real world, task identifiers are not always available. Therefore, class incremental learning (CIL) requires testing the mixed dataset $D_{test}^{1,...,t}$ of all samples in all learned tasks $\{T_1,...,T_t\}$:
\begin{equation}
    \label{c}
    \mathop{max}\limits_{\theta}E_{(x_j,y_j)\sim \{T_1,...,T_t\}}[logp_{\theta}(y_j|x_j)]
\end{equation}

\subsection{Spiking Neural Network} 
The spiking neural network, as a third generation neural network, has the advantages of high biological interpretability, efficient spatio-temporal joint information transfer, and low energy consumption~\cite{gerstner2002spiking,roy2019towards}. The most essential difference between the spiking neural network (SNN) and deep neural network (DNN) is that SNN uses spiking neurons instead of traditional artificial neurons. Spiking neurons receive and transmit  discrete spikes, aligning with the information transmission patterns of the biological brain and reducing energy consumption. We use the common leaky integrate-and-fire (LIF) spiking neuron~\cite{abbott1999lapicque}. The output spike $O_i^{t,l}$ of LIF neuron $i$ in layer $l$ at step $t$ is calculated as follows:

\begin{equation}
	\label{u}
	U_i^{t,l} =\tau (1-U_i^{t-1,l})+\sum_{j=1}^{M_{l-1}} W_{ij} O_j^{t,l-1}
\end{equation}
\begin{equation}
	\label{o}
	O_i^{t,l}=\left\{\begin{matrix}
		1, &  U_i^{t,l}\geq V_{th}\\ 
		0, &  U_i^{t,l}\textless V_{th}\\ 
	\end{matrix}\right.
\end{equation}
where $\tau=0.2$ is the time constant, $U_i^{t,l}$ is the membrane potential, and $V_{th}$ is the spiking threshold. Since discrete spike information hinders the gradient-based backpropagation algorithm, deep SNN uses surrogate gradients to optimize the network. Specifically, a derivable continuous function is used to approximate the derivative of the spike. The proposed algorithm uses Qgategrad~\cite{qin2020forward} surrogate gradients algorithm:
\begin{equation}
	\label{g}
    \frac{O_i^{t,l}}{U_i^{t,l}}=
        \begin{cases}
        0, & |U_i^{t,l}| > \frac{1}{\lambda} \\
        -\lambda ^2|U_i^{t,l}|+\lambda, & |U_i^{t,l}| \leq \frac{1}{\lambda}
        \end{cases}
\end{equation}

\begin{figure*}[!h] 
	\centering  
	\includegraphics[width=0.93\linewidth]{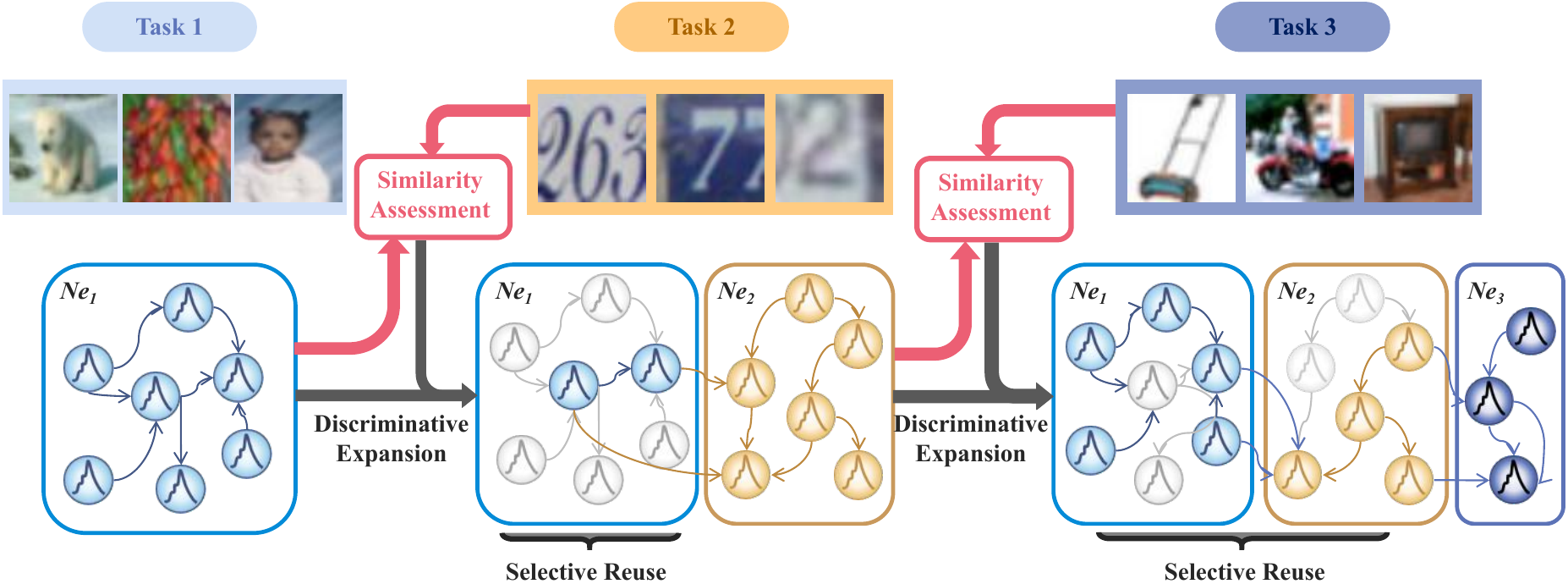} 
	\caption{The architecture of SCA-SNN. The SCA-SNN first evaluates the similarity between the new task and each old task, then discriminatively extends new neurons and selectively reuses learned neurons based on the degree of new task similarity. }
	\label{fig0}
\end{figure*}

\section{Method}
We elucidate the SCA-SNN algorithm as follows. An overview flow in Section \mbox{\ref{flow}}, the computational details in Section \mbox{\ref{det}}, including the similarity evaluation, discriminative expansion and selective reuse.

\subsection{SCA-SNN Architecture}
\label{flow}
\textbf{Similarity-based context assessment.} As shown in Fig. \ref{fig0}, context assessment serves as the first step of the proposed algorithm. When a new task arrives, SCA-SNN first evaluates the similarity $S_{t,p}, p=1,...,t-1$ between the new task $T_t$ and each of the previous tasks ${T_1,...,T_{t-1}}$ to guide subsequent expansion and reuse. This evaluation combines data similarity with the current network state, and does not require an additional evaluation network. A smaller value of $S_{t,p}$ indicates higher similarity between the two tasks. When learning the first task, SCA-SNN is trained as SNN routine in a predefined initial network. 

\textbf{Neuronal discriminative expansion.} Based on the above  similarity, to balance the cost of network expansion while considering the learning capacity of the new task, we compute the number of neuron expansions for the new task, which is proportional to $S_{t,p}$. When two tasks are similar, the network expansion scale dynamically decreases. Neuron population $Ne_{t}^{l}$ is added to the layer $l$, and $Ne_{t}^{l}$ is initially fully connected to all the neurons in the previous layer $l$-1. which contains all the existing neurons of the old task $\{Ne_{1}^{l-1},...,Ne_{t-1}^{l-1}\}$ and the newly expanded neurons $Ne_{t}^{l-1}$. Existing neurons of the old task can not add input synapses and change the weights of the input synapses but can add new learnable output synapses.

\textbf{Neuronal selective reuse.} During the training of the new task, SCA-SNN uses a gradient-based neuron relevance assessment method, combined with the similarity to select neurons from previous tasks that contribute to the new task $t$. The gradient-based relevance assessment excludes neurons that are irrelevant or conflicting with the new task, thereby reducing interference between tasks. The similarity value determines the number of reused neurons: the smaller the similarity $S_{t,p}$, the more beneficial neurons from the old tasks $Ne_{p}^{l}$ are reused. Thus, when the new task is similar to the old one, fewer neurons are extended, and more neurons of the old task are selectively reused, improving the learning ability of the new task while reducing energy consumption.

\subsection{SCA-SNN Computational Details}
\label{det}
In this section, we introduce the detailed scheme we use throughout this paper.

\subsubsection{Similarity-based context assessment}
When a new task arrives, SCA-SNN does not modify the network structure immediately. Instead, it first evaluates the relationship between the new task and each of the previous tasks. By alternating between similarity assessment and learning the new task within a unified network, SCA-SNN integrates data and the current network state without requiring additional similarity assessment networks.



In the continual learning scenario, we can only obtain the features of the current task. In order to judge the similarity between the current task and the previous task, we save the mean value of the sample features $M_{t,t}$ of each task category $t$ as the feature anchors after each task for subsequent task calling. The feature anchor size is the product of the number of categories per task and the feature dimension, which does not cause additional memory burden. Specifically, we input the new task samples into each of the existing old task neuron populations to extract features $F_{t,p}, p=1,...,t-1$ and compute the feature means $M_{t,p}$ for each category $c\in C_t$. To ensure the scatter stability in SNN, we introduce the  coefficient $\gamma<1$. It only guarantees that the scatter is greater than 0, and does not affect the relative magnitude of the similarity. In summary, according to the independent sampling KL divergence calculation method proposed by \mbox{\cite{wang2009divergence}}, the KL scatter similarity of current task $t$ and task $p$ is calculated as:

\begin{equation}
	\label{s}
	KL_{t,p}=\sum_{c\in C_t} \log \frac{||F_{t,p}-M_{p,p}||_2}{\gamma ||F_{t,p}-M_{t,p}||_2}
\end{equation}
Then we map the KL scatter to the similarity ranging from [0, 1] as the following formula:

\begin{equation}
	\label{s}
	S_{t,p}=\min \{ KL_{t,p},1-e^{2 KL_{t,p}}\}
\end{equation}
We obtain the similarity between the new task and each old task $\{S_{t,1},...,S_{t,t-1}\}$. 

\subsubsection{Neuronal discriminative expansion}
Depending on the difference in similarity-based context association between the current and previous tasks, SCA-SNN discriminatively expands different scale neuron populations joining the network for each new task to learn new features. The association magnitude $A$ of the new task with the previous task is defined as the minimum value of the similarity between the new task and each old task as follow:
\begin{equation}
	\label{a}
    A=\min \{S_{t,1},...,S_{t,t-1}\}
\end{equation}
A smaller value of $A$ indicates a stronger association, implying that many features already learned in the old tasks are similar to the new task and can be reused (see the next subsection). Therefore, the number of newly expanded neurons can be reduced without affecting the learning ability. To ensure that the number of expanded neurons remains within a reasonable range, we design the following adaptive calculation method to determine the number of neurons to be expanded in layer $l$:
\begin{equation}
	\label{n}
    NUM_t^l=M(1-e^{-\alpha A})
\end{equation}
where $M$ is the maximum number of neuron expansions, and the minimum value is 0.

\subsubsection{Neuronal selective reuse}
In the developmental habituation mechanism of the brain, neurons of learned habituation require repeated stimulation to re-fire, while irrelevant ones are disconnected~\cite{grissom2009habituation}. Inspired by this, we propose a gradient-based neuronal relatedness assessment to determine which neurons from the previous tasks are beneficial for the new task. At the beginning of the new task learning, the newly expanded neurons are fully connected to all the neurons of the old task. We learn the weights of the newly expanded neurons while keeping the old task neurons' input synaptic weights unchanged and calculating their gradients. Larger gradients indicate that the synaptic weights would require significant changes to be effective for the new task, suggesting that these synapses are not relevant to the new task and may hinder learning if left unchanged. In our SCA-SNN model, we evaluate relatedness at the neuron level by summing the input synaptic gradients $G_{p}^{l}$ of neurons from the old task.

As habituation activation in the brain requires repeated stimulation, the re-selection of neurons by the new task in our SCA-SNN also requires multiple judgments. We use a neuron relatedness function $R_{p}^{l}$ to judge that when the function of the old task neuron to the new task relatedness stays high multiple times, that neuron is re-selected by the new task network. Conversely, when the neuronal relatedness function is low for many times, the neuron will be judged as an irrelevant neuron excluded from the new task network. The relatedness function formula is as follow:
\begin{equation}
	\label{r}
	R_{p}^{l}=0.99 R_{p}^{l} -e^{-\frac{epoch}{2}}(2Norm(G_{p}^{l})-\rho_p)
\end{equation}
Where the negative exponent $e^{-epoch/2}$ make the pruning speed decreases with the epoch. This promotes the gradual stabilization of the network structure while knowledge learning is maturing, in line with the developmental process of biological brain networks that first rapidly decay and then gradually stabilize~\mbox{\cite{huttenlocher1979synaptic}}.

We disconnect the old task neurons with $R_{p}^{l}$ less than 0 from the new task neurons. Here, $\rho_p$ is determined by the similarity $S_{t,p}$ between the new and old tasks to which the current neuron belongs. The larger the $S_{t,p}$, the smaller $\rho_p$ is, and the more neurons are disconnected. $\rho_p$ is calculated as follow:
\begin{equation}
	\label{rp}
	\rho_p=\beta-S_{t,p}+bias
\end{equation}
where $bias$ is negatively correlated with layer $l$. In SCA-SNN, the new task is assessed with each previous old task for different similarities. When the new task $T_i$ is strongly similar to the old task $T_p$, more neurons from the old task $Ne_{p}^{l}$ are noticed by the new task and reactivated. Conversely, when the new task is weakly similar to the old task, more neurons of the old task are disconnected from the new task.

We present the specific procedure of our SCA-SNN algorithm as follows Algorithm \ref{alg2}.

 \begin{algorithm}[t]
	\caption{The SCA-SNN algorithm}
	\label{alg2}
	\KwIn{Dataset $D_{train}^t$ and $D_{test}^t$ for each task $t$.}
	\KwOut{Prediction class $y$.}
	\For{$t$ in sequential task $N$}{
        \eIf{t==0}
	{
		$S_{0,0}=0$;\\
        Initialize SNN for task 0;\\
	}
	{
        \%Similarity-based context assessment\\
        Evaluating new task similarities with old tasks $\{S_{t,1},...,S_{t,t-1}\}$ as Eq. \mbox{\ref{s}};\\
        \%Discriminative expansion\\
        Expanding new neurons for each SNN layer as Eq. \ref{a} and \ref{n};\\    
    }
		\For{$e$ in $Epoch$}{
	
	SNN forward prediction as Eq. \ref{u} and \ref{o};\\
	
	Backpropagation to update SNN as Eq. \ref{g}; \\

    \%Selective reuse\\
    Calculating the neuronal relatedness function $R_{p}^{l}$ as Eq. \ref{r} and \ref{rp};\\

    Prune irrelevant neurons with $R_{p}^{l} < 0$;
		}
	}
\end{algorithm}	

\section{Experimental Results}

\subsection{Datasets and Models}
To validate the effectiveness of the SCA-SNN model, we not only conduct extensive experiments on the generalized CIFAR100 and ImageNet (Mini-ImageNet and Tiny-ImageNet) datasets but also test the proposed similarity-based context aware algorithm in the mix-task datasets FMNIST-MNIST and SVHN-CIFAR100. The specific experimental datasets are as follows:

\begin{itemize}

    \item [$\bullet$] Split CIFAR100: We train the natural image dataset CIFAR100 in 10 steps split (10 new classes per step) and 20 steps split (5 new classes per step) without pre-training phase. 

    \item [$\bullet$] Split ImageNet: We randomly select 100 classes and 200 classes to form the Mini-ImageNet and Tiny-ImageNet datasets, and split them into 10 steps. 

    \item [$\bullet$] Permuted FMNIST - MNIST: We respectively permute the FMNIST and MNIST datasets to five tasks via random permutations of the pixels. Each task contains ten classes, divided into 60,000 training and 10,000 test samples. We alternately learn FMNIST and MNIST for a total of ten tasks.

    \item [$\bullet$] Rotated SVHN - Split CIFAR100: The SVHN dataset is arranged into five tasks by rotational transformation at different angles. We alternate between learning five Rotated SVHM tasks and five Spilt CIFAR100 tasks. Each task consists of 10 classes.
    
\end{itemize}

As for the SNN models, we are based on the Braincog Framework~\cite{zeng2023braincog}, and all use ResNet18 structure, except for the FMNIST-MNIST dataset using the model with two convolutional layers and two fully-connected layers.  For CIL, we set the total number of samples replayed to 2000 as~\cite{rebuffi2017icarl,rajasegaran2019random}. The time window $T$ is 4. Our code is available at \href{https://github.com/BrainCog-X/Brain-Cog/tree/main/examples/Structural_Development/SCA-SNN}{https://github.co m/BrainCog-X/Brain-Cog/tree/main/examples/Structural\_D evelopment/DSD-SNN}.

\begin{table*}[htbp]
  \centering
  \caption{The comparative performance of SCA-SNN on DNN-based continual learning for CIFAR100.}
    \begin{tabular}{l|c|c|c|c}
    \toprule
    \multicolumn{1}{c|}{\multirow{2}[4]{*}{\textbf{Method}}}  & \multicolumn{2}{c|}{\textbf{10 Steps }} & \multicolumn{2}{c}{\textbf{20 Steps }} \\
\cmidrule{2-5}      &  \textbf{TIL Accuracy (\%) } & \textbf{CIL Accuracy (\%)} & \textbf{TIL Accuracy (\%)} & \textbf{CIL Accuracy (\%)} \\
    \midrule
    EWC~\cite{kirkpatrick2017overcoming} & 61.11±1.43 & 7.25±0.09 & 50.04±4.26 & 4.63±0.04 \\
    MAS~\cite{aljundi2018memory} & 64.77±0.78 & 7.07±0.12 & 60.40±1.74 & 4.66±0.02 \\
    TAME~\cite{zhu2024tame} & 61.06 & - & 62.39 &  \\
    SI~\cite{zenke2017continual}  & 64.81±1.0 & 7.26±0.11 & 61.10+083 & 4.63±0.04 \\
    ERP~\cite{saha2023saliency}  & - & - & 60.08±0.35 & 13.70±0.84 \\
    OWM~\cite{zeng2019continual}  & 59.90±0.84 & 29.00±0.72 & 65.40±0.07 & 24.20±0.11 \\
    HNET~\cite{von2019continual}  & 63.57±1.03 & - & 70.48±0.25 & -\\
    LSTM\_NET~\cite{chandra2023continual}    & 66.61±3.77 & - & 79.96±0.26 & -\\
    CPG~\cite{hung2019compacting}& 70.15±3.95 & - & 82.60±0.30 & - \\
    PASS~\cite{zhu2021prototype} & 72.40±1.23 & 36.80±1.64 & 76.90±0.77 & 25.30±0.81 \\
    SGP~\cite{saha2023continual}  & 76.05±0.43 & - & 59.05±0.66 & - \\
    ERK~\cite{yildirim2024continual}  & 76.63±0.41 & - & 79.61±0.59 & -\\
    OSN~\cite{hutask2024task}  & 79.08±0.06 & - & 63.81±0.06 & -\\
    RDER~\cite{wang2024relational} & 80.96±0.34 & 48.62±0.48 & 81.79±0.52 & 41.01±0.86 \\
    Mnemonicst~\cite{liu2020mnemonics}& 82.30±0.30 & 51.00±0.34 & 86.20±0.46 & 47.60±0.47 \\
    HAT~\cite{serra2018overcoming}& 84.00±0.23 & 41.10±0.93 & 85.00±0.85 & 26.00±0.83 \\
    iCaRL~\cite{rebuffi2017icarl} & 84.20±1.04 & 51.40±0.99 & 85.70±0.68 & 47.80±0.48 \\
    DER++~\cite{buzzega2020dark}  & 84.20±0.47 & 55.30±0.10 & 86.60±0.50 & 46.60±1.44 \\
    \midrule
    \textbf{ Our SCA-SNN} & \textbf{85.61±0.24} & \textbf{57.06±0.29} & 
    \textbf{86.45±0.35} & \textbf{50.19±0.45} \\ 
    \bottomrule
    \end{tabular}%
  \label{tab1}%
\end{table*}%

\subsection{Comparison with SNN-based continual learning}
\begin{figure}[t] 
	\centering  
	\includegraphics[width=1\linewidth]{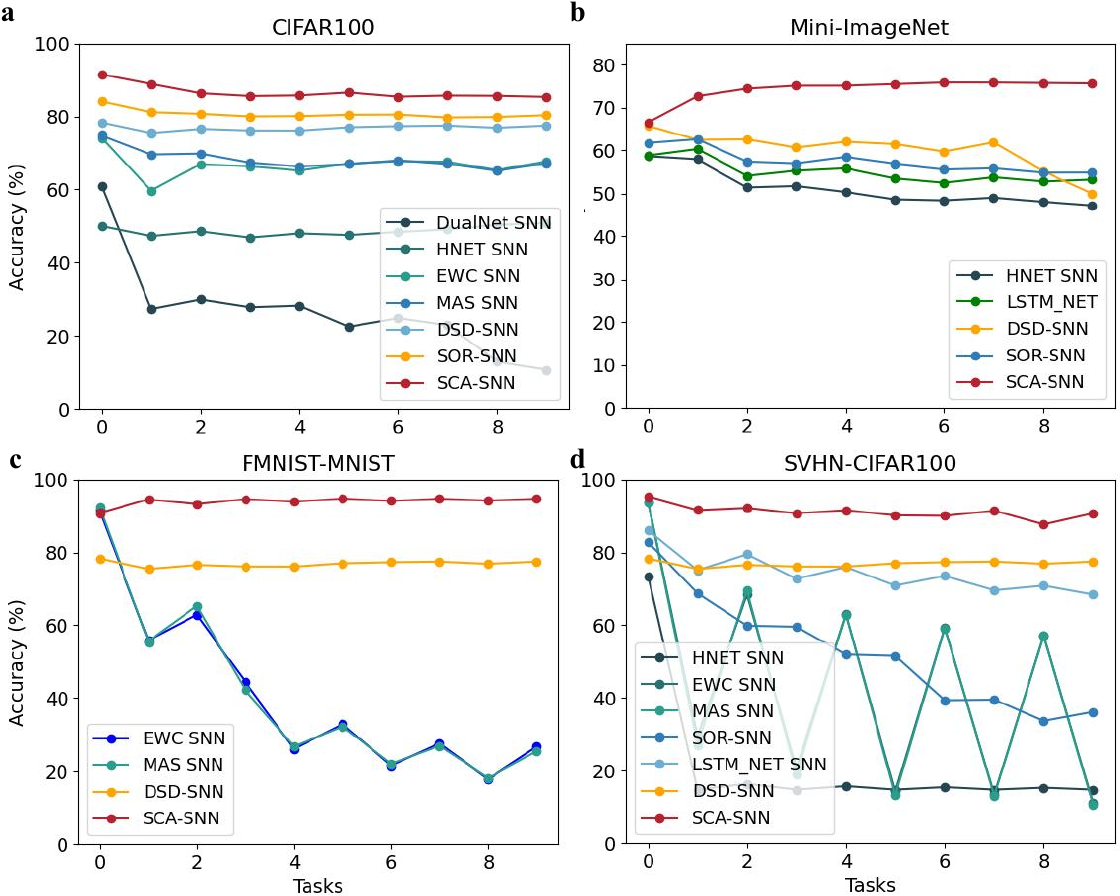} 
	\caption{The performance comparison of SCA-SNN on SNN-based continual learning in task incremental learning.}
	\label{fig1}
\end{figure}

First, we compare the few existing deep SNN-based structure-expanded continual learning algorithms DSD-SNN~\mbox{\cite{han2023enhancing}} and SOR-SNN~\mbox{\cite{han2023adaptive}}. In addition, since SNN-based continual learning algorithms belong to the preliminary exploration with less existing work, we reproduce some DNN-based continual learning in SNNs for comparison to show the superiority of our algorithms in SNNs, 
 including  EWC\mbox{~\cite{kirkpatrick2017overcoming}}, MAS~\mbox{\cite{aljundi2018memory}}, DualNET~\mbox{\cite{pham2021dualnet}}, HNET~\mbox{\cite{von2019continual}} and LSTM$\_$NET ~\cite{chandra2023continual}. As shown in Fig. \ref{fig1}, our SCA-SNN consistently demonstrates superior performance throughout the learning process for all tasks in both general and mixed datasets. Particularly in generalized high-resolution Mini-ImageNet, the SCA-SNN model achieves a significant improvement in the accuracy. Our average accuracy is 75.7\%, which is 20.7\% higher than SOR-SNN, with the second highest accuracy. Moreover, compared with the decreasing accuracy trend of other methods, the accuracy of our SCA-SNN is stable at around 75\%, which indicates that proper selective reuse not only efficiently utilizes the past learned knowledge to help learn new tasks but also eliminates the interference of irrelevant knowledge on new tasks.

In the mixed dataset, the accuracy of our SCA-SNN remains stable across tasks of different types. However, the EWC, MAS, LSTM$\_$NET, and SOR-SNN methods all exhibit task-related regularity fluctuations during the alternating learning process of mixed datasets. These methods perform well only on a set of similar tasks, with a significant decrease in accuracy on the other set of tasks. In particular, two regularization methods, EWC and MAS, fluctuate dramatically. For example, in the SVHN-CIFAR100 dataset, when the first task is from the SVHN dataset, these two methods are effective only for SVHN tasks, while their accuracy on CIFAR100 tasks approaches random selection. HNET completely loses its learning ability when the task type changes. This indicates that the flexibility of these algorithms in responding to different tasks in SNNs requires enhancement. Continual learning algorithms specifically designed for SNNs that can adapt to dynamically changing environments are urgently needed. Unlike the above methods, our SOR-SNN achieves an average accuracy of 90.73\% when the first task comes from SVHN, which is higher than the second most accurate one, DSD-SNN, by 13.25\%. When the first task is from the CIFAR100 dataset, SCA-SNN achieves an average accuracy of 90.31\%, which is not significantly different from the performance before the exchange of tasks. This is attributed to our similarity-based context aware algorithm, which effectively recognizes the similarity between tasks and performs targeted neuron expansion and reuse, allowing the network to adapt effectively to different tasks.

\begin{table}[htbp]
  \centering
  \caption{The comparative performance of SCA-SNN on DNN-based continual learning for 10steps Tiny-ImageNet.}
    \resizebox{3.3in}{!}{\begin{tabular}{l|cc}
    \toprule
    \textbf{Method} & \textbf{TIL Acc(\%)} & \textbf{CIL Acc(\%)} \\
    \midrule
        HNET~\cite{von2019continual} & 27.8±0.86 & 5.8±0.56 \\
    OWM~\cite{zeng2019continual} & 28.1±0.55 & 8.6±0.42 \\   
    MUC~\cite{liu2020more} & 47.2±0.22 & 17.4±0.17 \\
    PASS~\cite{zhu2021prototype} & 47.6±0.38 & 18.7±0.58 \\
    BiC~\cite{wu2019large}  & 50.3±0.65 & 21.2±0.46 \\
    Mnemonicst~\cite{liu2020mnemonics} & 52.9±0.66 & 28.5±0.72 \\
    LwF~\cite{li2017learning}& 55.3±0.35 & 24.3±0.26 \\
    DER++~\cite{buzzega2020dark}  & 59.7±0.60 & 30.5±0.30 \\
    HAT~\cite{serra2018overcoming} & 63.8±0.41 & 29.8±0.65 \\
    SupSup~\cite{wortsman2020supermasks} & 64.4±0.20 & 27.0±0.45 \\
    RDER~\cite{wang2024relational}&68.8 ±0.54&39.7 ±0.96 \\
    \midrule
    \textbf{Our SCA-SNN} & \textbf{71.92±0.52} & \textbf{42.35±0.59} \\
    \bottomrule
    \end{tabular}
  \label{tab2}}
\end{table}%

\subsection{Comparison with DNN-based continual learning}
For extensive comparison and analysis, we compare our method with mature DNN-based continual learning methods on task incremental and class incremental learning. For the CIFAR100 dataset as Tab. \mbox{\ref{tab1}}, our method achieves an accuracy of 85.61±0.24\% in TIL and 57.06.±0.29\% in CIL of 10steps. Moreover, in the 20steps scenario with more tasks, SCA-SNN achieves superior accuracy of 86.45±0.35\% in TIL and 50.19±0.45\% in CIL, respectively. These results outperform the structural expansion algorithms such as ERK~\mbox{\cite{yildirim2024continual}}, with improvements of 8.98\% and 8.64\% in TIL for the 10steps and 20steps scenarios, respectively, and outperform HAT~\mbox{\cite{serra2018overcoming}} by 15.96\% and 24.19\% in CIL. For the more complex Tiny-ImageNet dataset in 10steps scenario, as shown in Tab. \mbox{\ref{tab2}}, our algorithm achieves the highest performance among a series of DNN-based continual learning algorithms. The average accuracy of SCA-SNN reaches 71.92±0.52\%, which improves accuracy by 3.10\% compared to the RDER~\cite{wang2024relational} algorithm, whose accuracy is the next highest. In summary, although our SCA-SNN algorithm is implemented based on SNN, we achieve the same favorable performance compared to DNN-based algorithms.

\begin{table*}[h]
    \centering
    \caption{Energy consumption comparisons for 10steps CIFAR100.}
    \resizebox{5.5in}{!}{
\begin{tabular}{lccccc}
    \toprule
    \toprule
    \multicolumn{1}{c}{\textbf{Method}} & \makecell{\textbf{Memory} \\ \textbf{ Method}} &  \makecell{\textbf{Number of} \\ \textbf{Connections}} & \makecell{\textbf{Number of} \\ \textbf{ Neurons}} & \textbf{FLOPs} & \makecell{\textbf{Computational} \\ \textbf{ Energy}} \\
    \midrule
    EWC SNN & Regularization & 11.2$ \times10^6$ & 3840 & 11.1$ \times10^8$ & 4.0$ \times10^9$ pJ \\
    MAC SNN & Regularization & 11.2$ \times10^6$ & 3840 & 11.1$ \times10^8$ & 4.0$ \times10^9$ pJ \\
    PODNet DNN~\cite{douillard2020podnet} & Regularization & 11.2$ \times10^6$ & 3840 & 11.1$ \times10^8$ & 5.1$ \times10^9$ pJ \\
    iCaRL DNN~\cite{rebuffi2017icarl} & Replay & 11.2$ \times10^6$ & 3840 & 11.1$ \times10^8$ & 5.1$ \times10^9$ pJ \\
    DER++ DNN~\cite{buzzega2020dark} & Replay & 11.2$ \times10^6$ & 3840 & 11.1$ \times10^8$ & 5.1$ \times10^9$ pJ \\
    DSD-SNN~\cite{han2023enhancing} & expansion & 115.8$ \times10^6$ &  5362 & 38.4$ \times10^8$  & 13.8$ \times10^9$ pJ \\
    DER DNN~\cite{yan2021dynamically} & expansion & 61.6$ \times10^6$ & 21472 & 61.1$ \times10^8$ & 28.1$ \times10^9$ pJ \\
    \textbf{Our SCA-SNN} & expansion & 8.4$ \times10^6$ & 3111 & 9.2$ \times10^8$ & 3.3$ \times10^9$ pJ \\
    \bottomrule
    \bottomrule
    \end{tabular}}%
		\label{tab3}%
	\end{table*}%

\subsection{Energy consumption}

We compare the network energy consumption of the proposed methods in terms of number of connections, number of neurons, floating point operations per second (FLOPs), and computational energy, as Tab. \mbox{\ref{tab3}} shows. Among them, the number of connections and neurons is the average of the number of weights and neurons activated by each task. The computational energy follows the widely used ~\mbox{\cite{chakraborty2021fully}}, for DNN: 
\begin{equation}
	\label{es}
	E_{SNN}=FLOPS_{SNN}*E_{MAC}
\end{equation}
where $E_{MAC}=4.6pJ$ is the energy consumption of multiply-accumulate (MAC) operations. For SNN: 
\begin{equation}
	\label{es}
	E_{SNN}=FLOPS_{SNN}*E_{AC}*T
\end{equation}
where $E_{AC}=0.9pJ$ is the energy consumption of accumulate (AC) operations.

The results show that after completing the learning of all tasks, the parameter pruning rate of our network is 42.08\%, and the average number of connections and neurons used in testing are respectively 8.4$\times10^6$ and 3111, which are significantly lower than the other methods. In this case, the required FLOPs are also smaller. This is because when learning similar tasks, our SCA-SNN model reuses more existing neurons to reduce the expansion of new neurons. When learning dissimilar tasks, the existing irrelevant neurons are heavily pruned.
Therefore, compared to regularization and replay methods, our actual number of activated connections and neurons is less than the standard structure of ResNet18. In comparison to structure expansion algorithms, our similarity-based discriminative expansion and selective reuse increase the sparsity of reused neurons while reducing the number of newly expanded neurons.

In addition, compared to DNN, our SNN-based method replaces the multiply-accumulate operation with the accumulate operation. Under the same FLOPs, SNN-based methods require lower computational energy. Since our method has lower FLOPs and is based on SNNs, it achieves the lowest computational energy 3.3$ \times10^9$ pJ. 
 In summary, with the dual effect of the sparsity of the proposed SCA method and the high efficiency of SNN, our SCA-SNN method significantly reduces the network energy consumption.
\begin{figure}[!h] 
	\centering  
	\includegraphics[width=1\linewidth]{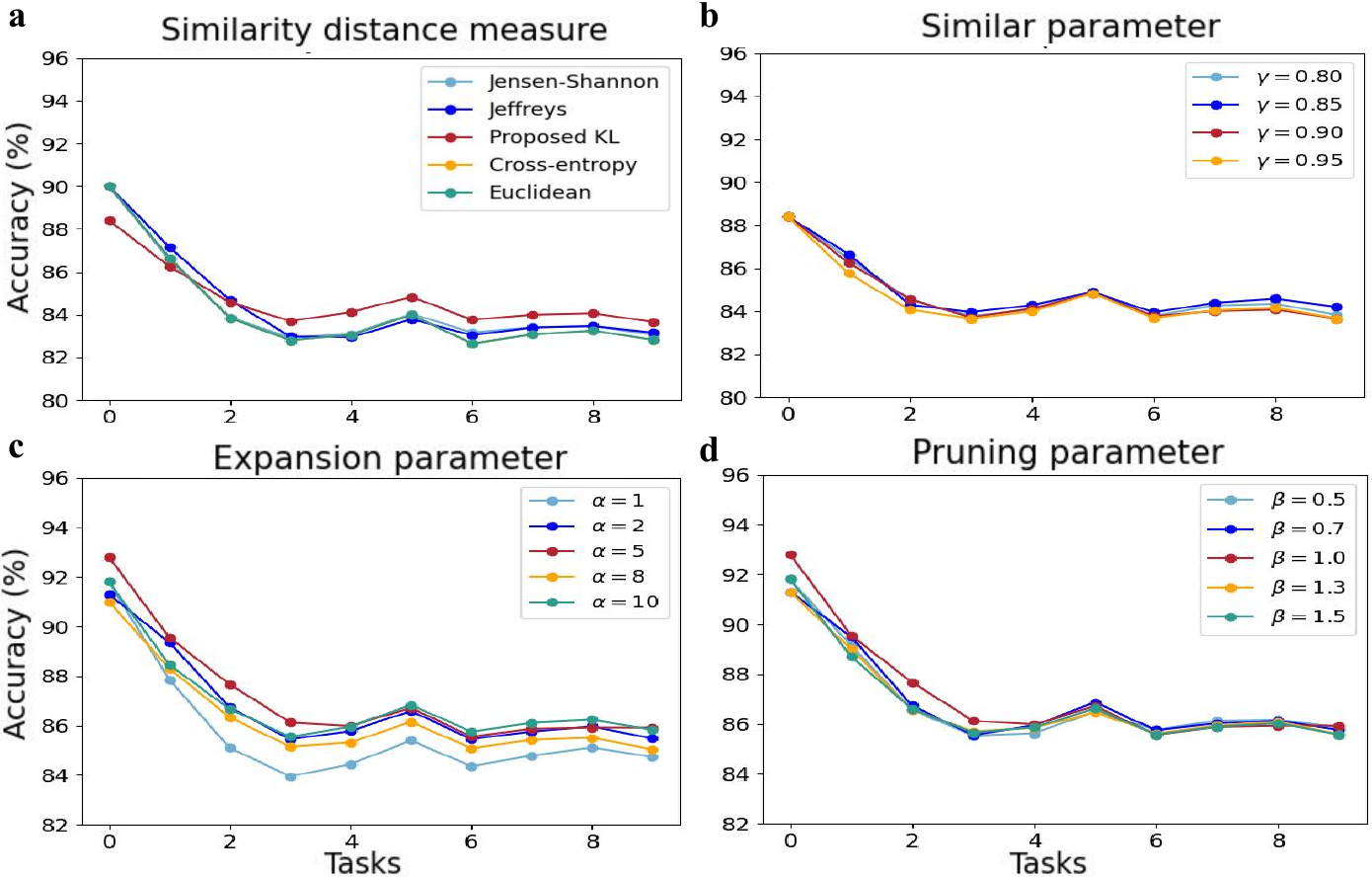} 
	\caption{The effect on the performance of different similarity assessments (\textbf{a})  and the parameters of proposed similarity (\textbf{b}), neuronal discriminative expansion (\textbf{c}) and selective reuse (\textbf{d}).}
	\label{fig2}
\end{figure}

\subsection{Ablation Studies}

We validate the effectiveness of the proposed similarity assessment method and the parameter robustness of the proposed SCA-SNN, including: similarity parameter $\gamma$, expansion parameter $\alpha$, and reuse parameter $\beta$. For similarity assessment, we compare the proposed KL-based algorithm with other commonly used methods for similarity distance assessment, including Jensen-Shannon divergence, Jeffreys divergence, Cross-entropy and Euclidean distance. The results in Fig. \mbox{\ref{fig2}} a show that the network evaluating similarity with the proposed KL-based method is slightly higher in accuracy than the other methods. This demonstrates the advantage of KL-based similarity assessment in supporting the discriminative extension and selective reuse mechanisms to achieve flexible and efficient continual learning in SNNs. In addition, we verify the robustness of the parameter $\gamma$ in the proposed KL-based similarity, as shown in Fig. \mbox{\ref{fig2}} b. The results show that the accuracy is almost unchanged under different $\gamma$, which demonstrates again that the $\gamma$ serves to ensure the non-negativity of the similarity without affecting the performance.

For the effect of neuron discriminative expansion parameter $\alpha$ on the average accuracy of continual learning, the larger $\alpha$, the more number of extended neurons, and the higher the network energy consumption under the condition of a certain distance between tasks. As shown in Fig. \mbox{\ref{fig2}} c, the average accuracy curve of SCA-SNN at $\alpha$ = 5 is slightly higher than others. When $\alpha$ is smaller than 5, the smaller network scale leads to a minor decrease in the accuracy; when $\alpha$ is larger than 5, the larger network scale leads to the inability to reach the optimal value with a definite training cost. Nonetheless, the final accuracy of our SCA-SNN is stable at around 85\% with the variation of $\alpha$, demonstrating our model's robustness to the extended parameter $\alpha$. For the selective reuse parameter $\beta$, the larger $\beta$ is, the smaller the pruning rate of the network is. As shown in Fig. \mbox{\ref{fig2}} d, the performance of SCA-SNN remains roughly stable with varying $\beta$. When $\beta$=1, SCA-SNN achieves the highest performance of 85.92\%, at which the sparsity of the network is 42.08\%. The above indicates that our SCA-SNN is highly robust and stable to hyperparameters.
\begin{figure}[t] 
	\centering  
	\includegraphics[width=1\linewidth]{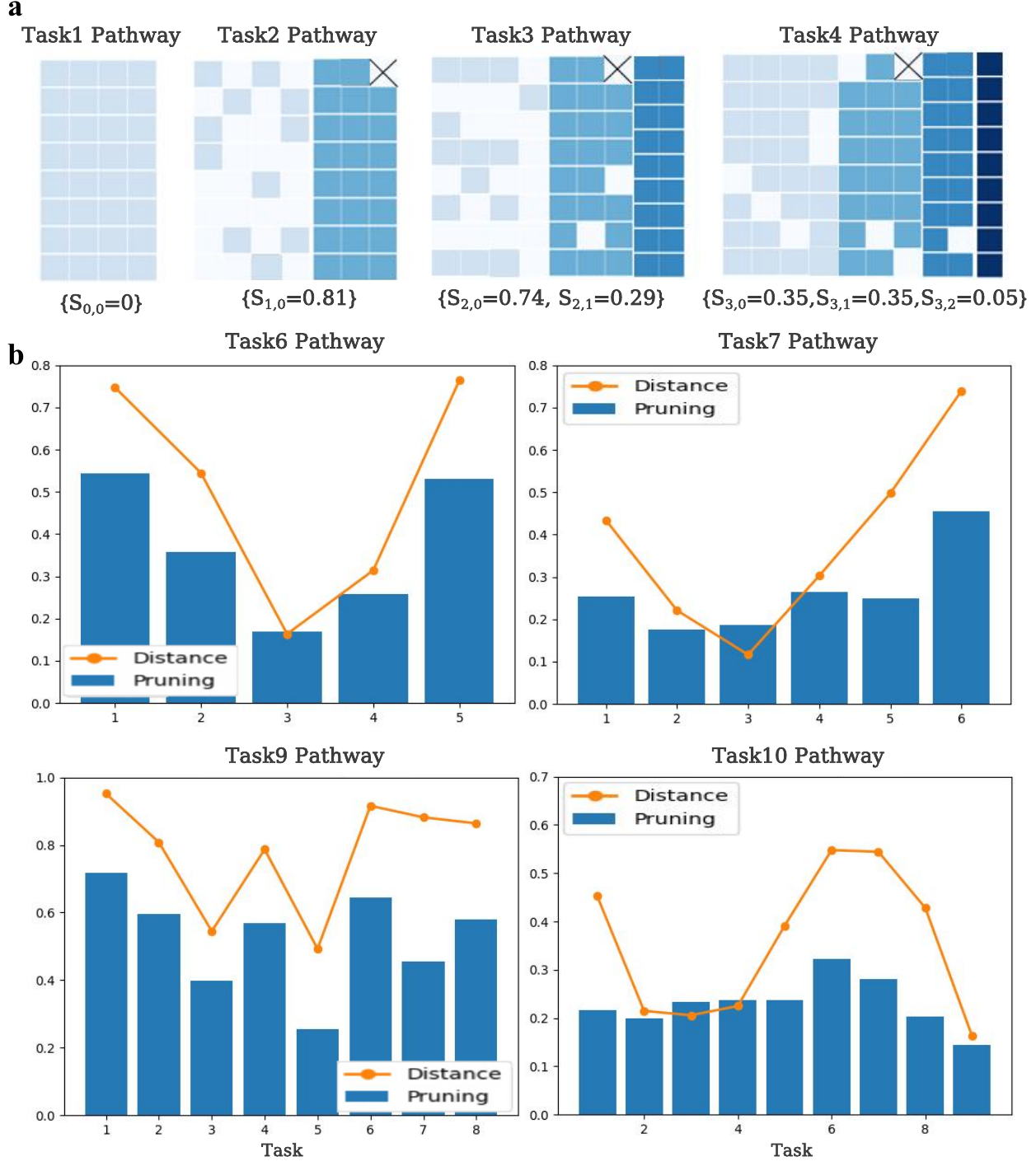} 
	\caption{Similarity-based neuronal population assignment. \textbf{a):} Colored blocks represent activated convolutional feature maps, and white blocks represent pruned feature maps. The fork blocks are placeholders only. \textbf{b):} The relationship between task-to-task distance and the pruning rate of new task pathways on neurons belonging to old tasks.}
	\label{fig3}
\end{figure}

\subsection{Similarity-based neuronal population assignment}
In the biological brain, the nervous system activates similar neuron populations when learning similar tasks to efficiently coordinate multiple cognitive tasks~\cite{sakurai1996hippocampal,sakurai1996population}. Our SCA-SNN learning process involves reusing neurons from past similar tasks and heavily pruning irrelevant neurons from past dissimilar tasks. The magnitude of newly expanded neurons is proportional to the similarity distance. Thus, when learning a new task $t$ similar to the previous task $t-1$, the neurons activated by both are more similar; when learning a new task unrelated to the previous task, the overlap of the neuron populations activated by both is less. Fig. \ref{fig3} a visualizes the expansion and reuse of the fourth convolutional layer of ResNet18 during the learning of Task 1 to Task 4 for the CIFAR100 10steps TIL. Task 1 initializes a dense network. As an example the distances of Task 3 from Tasks 1 and 2 are 0.74 and 0.29 respectively, indicating that Task 3 is more similar to Task 2 and less similar to Task 1. Thus task 3 heavily reuses neurons from task 2 and extends fewer neurons, while pruning more neurons already in task 1 that are irrelevant to task 3.

In addition, we statistics the relationship between similarity and the number of selectively reused neurons, the relationship between the distance magnitude among tasks and the pruning rate of task-related neurons. In Fig. \mbox{\ref{fig3}} b, it is observed that in the ResNet18 SNNs pathway for tasks 6, 7, 9, and 10, a larger distance indicates less similarity between tasks and leads to a higher pruning rate of neuron population belonging to the old task. For example, when learning the new task 7, the distance between task 7 and task 3 was evaluated to be 0.12, and the task 7 network pruned only 18.57\% of the neurons extended from task 3. The distance between task 7 and task 6 was evaluated to be a larger 0.74. Thus, the task 7 pathway pruned 45.46\% of the neurons belonging to task 6. This demonstrates that our SCA-SNN model can adaptively coordinate similar neuron populations for similar tasks through similarity-based discriminative expansion and selective reuse, aligning with the multi-task continual learning mechanism of the biological brain and improving learning efficiency.

\section{Discussion}
In the lifelong learning process of humans, the brain gradually acquires proficiency in various tasks and adapts to ever-changing environments. Among these processes, the task context recognition mechanism plays a crucial role by associating new tasks with similar previously learned tasks and selecting useful knowledge from the old tasks to facilitate the learning of new ones~\mbox{\cite{bar2007proactive,bar2004visual}}. Inspired by this, we propose the similarity-based context aware continual learning algorithm called SCA-SNN, which is based on the brain-like SNN infrastructure. When a new task arrives, SCA-SNN first uses its own network to identify the similarity relationships between the new task and each of the learned tasks. This information guides subsequent neuron expansion and reuse, ultimately leading to the adaptive allocation of similar neuron groups for similar tasks.

Unlike other continual learning algorithms, when faced with non-smooth continual learning tasks with large knowledge spans, the proposed algorithm associates tasks containing similar knowledge with similar neuron groups, assigning different tasks with their own neuron groups. Our method enables stable learning for different tasks while reducing interference and facilitating appropriate beneficial knowledge transfer. Thus, during the cross-continual learning process of different tasks, the accuracy of proposed algorithm continues to rise steadily. In contrast, regularization-based methods such as EWC SNN~\mbox{\cite{kirkpatrick2017overcoming}} and MAC SNN~\mbox{\cite{aljundi2018memory}} suffer from incompatibility between tasks due to significant differences in important synapses for different tasks, leading to a loss of continual learning ability after significant task-related fluctuations. Structure-based methods like SOR-SNN~\mbox{\cite{han2023adaptive}} and LSTM\_NET SNN~\mbox{
\cite{chandra2023continual}}, which independently select subnetworks for each task without considering the similarity relationships between tasks, also result in task-related performance fluctuations and a decline in final performance. Furthermore, by expanding and reusing neuron groups based on similarity relationships, our method avoids redundant neurons between similar tasks and interference between neurons of dissimilar tasks, significantly reducing energy consumption while effectively enhancing performance. Combined with the energy-saving attributes of discrete SNN networks, the proposed method reduces energy consumption by 1.21, 1.55, and 8.52 times, respectively, compared to the DNN-based regularization algorithm MAS~\mbox{\cite{aljundi2018memory}}, the replay algorithm iCaRL~\mbox{\cite{rebuffi2017icarl}}, and the structural expansion algorithm DER~\mbox{\cite{yan2021dynamically}}.

To verify the effectiveness of the proposed algorithm, we conducted extensive experiments not only on general datasets such as CIFAR100 and Tiny-ImageNet but also validated the performance of SCA-SNN in more realistic non-smooth cross-task learning scenarios. The experimental results demonstrate that the proposed algorithm exhibits stable and efficient learning capabilities for different tasks, with enhanced performance robustness, structural adaptability, and biological plausibility. However, SCA-SNN focuses on completing different image classification tasks and does not yet possess the comprehensive recognition and continual learning abilities for various cognitive tasks such as perception, decision-making, and reasoning. In the future, we will further enhance the degree of freedom in neuron allocation to achieve adaptive continual learning for dynamic tasks in real-world environments within a single efficient non-traditional structural network.

\section{Conclusion}
Inspired by brain contextual association mechanism, we propose a similarity-based context aware spiking neural network continual learning algorithm applied in both task incremental learning and class incremental learning. The proposed similarity assessment algorithm can effectively identify the association of different task contexts, guiding deep SNNs to perform neuron discriminative expansion and selective reuse, reducing task interference, and improving knowledge utilization efficiency. Our SCA-SNN algorithm achieves superior performance and reduces network energy consumption compared to both SNN-based and DNN-based continual learning for generalized and mixed datasets. Meanwhile, experiments demonstrate that our algorithm enables adaptive allocation of similar neuron populations for similar tasks like the biological brain.

\section{Acknowledgments}
This work is supported by the Strategic Priority Research Program of the Chinese Academy of Sciences (Grant No. XDB1010302), the National Natural Science Foundation of China (Grant No. 62106261) and the funding from Institute of Automation, Chinese Academy of Sciences (Grant No. E411230101)

\subsection{Author contributions}
B.Han, F.Zhao, and Y.Zeng designed the study. B.Han, F.Zhao, Y.Li and X.Li performed the experiments and the analyses. B.Han, F.Zhao, Q.Kong and Y.Zeng wrote the paper.

\subsection{Declaration of interests}
The authors declare that they have no competing interests.

\bibliographystyle{cas-model2-names}
\bibliography{mybibfile}

\end{document}